\pdfoutput=1
\documentclass[11pt]{article}

\usepackage[]{emnlp2021}

\usepackage{times}
\usepackage{verbatim}
\usepackage{subfigure}
\usepackage[T1]{fontenc}
\usepackage[utf8]{inputenc}

\usepackage{microtype}
\usepackage{xspace}
\usepackage{paralist}

\newcommand{\gyafc}{\textsc{gyafc}\xspace}
\newcommand{\xformal}{\textsc{xformal}\xspace}

\newcommand{\bert}{\textsc{bert}\xspace}
\newcommand{\bleu}{\textsc{bleu}\xspace}

\usepackage{tabu}

\newcommand{\reg}{\textsc{reg}\xspace}
\newcommand{\ppl}{\textsc{ppl}\xspace}

\newcommand{\cnn}{\textsc{cnn}\xspace}
\newcommand{\rob}{Ro\textsc{bert}a\xspace}
\newcommand{\gru}{\textsc{gru}\xspace}
\newcommand{\lstm}{\textsc{lstm}\xspace}

\newcommand{\sts}{\textsc{sts}\xspace}

\newcommand{\kenlm}{Ken\textsc{lm}\xspace}
\newcommand{\gpt}{\textsc{gpt}\xspace}

\newcommand{\nlp}{\textsc{nlp}\xspace}
\newcommand{\fost}{\textsc{f}o\textsc{st}\xspace}
\newcommand{\italian}{\textsc{it}\xspace}
\newcommand{\french}{\textsc{fr}\xspace}
\newcommand{\portuguese}{\textsc{br-pt}\xspace}
\newcommand{\english}{\textsc{en}\xspace}

\newcommand{\st}{\textsc{st}\xspace}
\newcommand{\cmark}{{\color{darkgreen}{\ding{51}}}}%
\newcommand{\xmark}{{\color{red}{\ding{55}}}}%

\usepackage{times}
\usepackage{latexsym}

\usepackage{amssymb}% http://ctan.org/pkg/amssymb
\usepackage{pifont}
\usepackage{graphicx}
\usepackage{booktabs} % To thicken table lines

\usepackage{colortbl}
\usepackage{multirow}
\usepackage{import}
\usepackage{cmll}
\usepackage{MnSymbol}

\newcommand*\rot{\rotatebox{90}}

\usepackage{arydshln}
\definecolor{red}{rgb}{0.3, 0.2, 0.3}
\definecolor{blue}{rgb}{0., 0.3, 0.9}
\definecolor{whitesmoke}{rgb}{0.91,0.91,0.91}
\definecolor{whitesmoke}{rgb}{0.93,0.93,0.93}
\definecolor{rr}{rgb}{0.8, 0.2, 0.3}

\definecolor{darkgreen}{rgb}{0.0, 0.42, 0.24}

\newcommand{\C}{\cellcolor{whitesmoke}}

\newcommand{\sterp}{\textsuperscript{\includegraphics[scale=0.02]{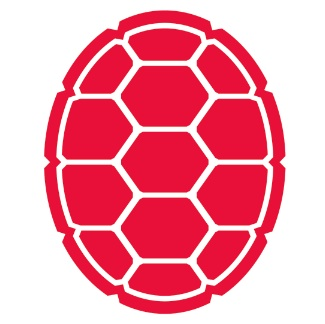}}}

\newcommand{\sdataminr}{\textsuperscript{\includegraphics[scale=0.02]{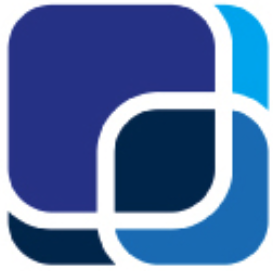}}}

\usepackage{colortbl}
\usepackage{pgfplots}
\usepackage{pgfplotstable}
    \usetikzlibrary{pgfplots.colorbrewer}
    \pgfplotsset{
        colormap/RdBu,
    }
    \pgfplotstableset{
        /color cells/min/.initial=0,
        /color cells/max/.initial=1000,
        /color cells/textcolor/.initial=,
        color cells/.code={%
            \pgfqkeys{/color cells}{#1}%
            \pgfkeysalso{%
                postproc cell content/.code={%
                    \begingroup
                    \pgfkeysgetvalue{/pgfplots/table/@preprocessed cell content}\value
                    \ifx\value\empty
                        \endgroup
                    \else
                    \pgfmathfloatparsenumber{\value}%
                    \pgfmathfloattofixed{\pgfmathresult}%
                    \let\value=\pgfmathresult
                    \pgfplotscolormapaccess
                        [\pgfkeysvalueof{/color cells/min}:\pgfkeysvalueof{/color cells/max}]
                        {\value}
                        {\pgfkeysvalueof{/pgfplots/colormap name}}%
                    \pgfkeysgetvalue{/pgfplots/table/@cell content}\typesetvalue
                    \pgfkeysgetvalue{/color cells/textcolor}\textcolorvalue
                    \toks0=\expandafter{\typesetvalue}%
                    \xdef\temp{%
                        \noexpand\pgfkeysalso{%
                            @cell content={%
                                \noexpand\cellcolor[rgb]{\pgfmathresult}%
                                \noexpand\definecolor{mapped color}{rgb}{\pgfmathresult}%
                                \ifx\textcolorvalue\empty
                                \else
                                    \noexpand\color{\textcolorvalue}%
                                \fi
                                \the\toks0 %
                            }%
                        }%
                    }%
                    \endgroup
                    \temp
                    \fi
                }%
            }%
        }
    }
\usepackage{xcolor}

\newcommand{\ignore}[1]{}

\title{Evaluating the Evaluation Metrics for Style Transfer: \\  A Case Study in Multilingual Formality Transfer}

\author{Eleftheria Briakou\sterp, Sweta Agrawal\sterp,  Joel Tetreault\sdataminr, Marine Carpuat\sterp  \\
  \sterp University of Maryland,
  \sdataminr Dataminr, Inc. \\
 \texttt{\href{mailto:ebriakou@cs.umd.edu}{ebriakou@cs.umd.edu}},
 \texttt{\href{mailto:sweagraw@cs.umd.edu}{sweagraw@cs.umd.edu}},\\
 \texttt{\href{mailto:jtetreault@dataminr.com}{jtetreault@dataminr.com}},
  \texttt{\href{mailto:marine@cs.umd.edu}{marine@cs.umd.edu}}

 }

\pgfplotsset{compat=1.17}
\begin{document}
\maketitle

\begin{abstract}
While the field of style transfer (\st) has been growing rapidly, it has been hampered by a lack of standardized practices for automatic evaluation.  
In this paper, we evaluate leading \st automatic metrics on the oft-researched task of formality style transfer.  Unlike previous evaluations, which focus solely on English, we expand our focus to Brazilian-Portuguese, French, and Italian, making this work the first multilingual evaluation of metrics in \st. 
We outline best practices for automatic evaluation in (formality) style transfer and identify several models that correlate well with human judgments and are robust across languages.  We hope that this work will help accelerate development in \st, where human evaluation is often challenging to collect.
\end{abstract}

%%% Section 1: Introduction
\section{Introduction}

Textual style transfer (\st) is defined as a generation task where a text sequence is paraphrased while controlling one aspect of its style~\cite{jin2021deep}.
For instance, the informal sentence in Italian \textit{``in bocca al lupo!''} (i.e., ``good luck'') is rewritten to the formal version \textit{``Ti rivolgo un sincero augurio!''} (i.e., ``I send you a sincere wish!'').
Despite the growing attention on \st in the \nlp literature \citep{jin2021deep}, progress is hampered by a lack of standardized and reliable automatic evaluation metrics. 
Standardizing the latter would allow for quicker development of new methods and comparison to prior art without relying on time and cost-intensive human evaluation that is currently employed by more than $70\%$ of \st papers~\cite{briakou-etal-2021-review}.

\st is usually evaluated across three dimensions: style transfer (i.e., has the style of the generated output changed as intended?), meaning preservation (i.e., are the semantics of the input preserved?), and fluency (i.e., is the output well-formed?). As we will see, a wide range of automatic evaluation metrics and models has been used to quantify each of these dimensions. 
%Concretely, prior work employs $9$, $5$, and $8$ different automatic metrics for formality, meaning, and fluency, respectively.
For example, prior work has employed as many as nine different automatic systems to rate formality alone (see Table~\ref{tab:status}).
However, it is not clear how different automatic metrics compare to each other and how well they agree with human judgments.
Furthermore, previous studies of automatic evaluation have exclusively focused on the English language  ~\cite{yamshchikov2020styletransfer,pang-2019-towards, Pang2019UnsupervisedEM, tikhonov-etal-2019-style, mir-etal-2019-evaluating}; yet, \st requires evaluation methods that generalize reliably beyond English. 

We address these limitations by conducting a controlled empirical comparison of commonly used automatic evaluation metrics. Concretely, for all three evaluation dimensions, we compile a list of different automatic evaluation approaches used in prior \st work and study
how well they correlate with human judgments. We choose to build on available resources as collecting human judgments across the evaluation dimensions is a costly process that requires recruiting fluent speakers in each language addressed in evaluation. While there are many stylistic transformations in \st, we conduct our study through the lens of formality style transfer (\fost), which is one of the most popular style dimensions considered by past \st work \citep{jin2021deep,briakou-etal-2021-review} and for which reference outputs and human judgments are available for four languages: English, Brazilian-Portuguese, French, and Italian.
\begin{itemize}
    \item We contribute a meta-evaluation study that is not only the first \textit{large-scale} comparison of automatic metrics for \st but is also the first work to investigate the robustness of these metrics in \textit{multilingual} settings.
    \item We show that automatic evaluation approaches based on a formality regression model fine-tuned on \textsc{xlm-r} and the chr\textsc{f} metric correlate well with human judgments for style transfer and meaning preservation, respectively, and propose that the field adopts their usage.  These metrics are shown to work well \textit{across} languages, and not just in English.
    \item We show that framing style transfer evaluation as a binary classification task is problematic and propose that the field treats it as a regression task to better mirror human evaluation.
    \item Our analysis code and meta-evaluation files with system outputs are made public to facilitate further work in developing better automatic metrics for \st: \url{https://github.com/Elbria/xformal-FoST-meta}.  
\end{itemize}

%%% Section 2: Current Status
\section{Background}
\label{sec:status}
%
%%%% Status table
\renewcommand{\arraystretch}{1.02}
\begin{table*}[!t]
    \centering
    \scalebox{0.65}{
    \begin{tabular}{l@{\hskip 0.7in}llr@{\hskip 0.6in}lll@{\hskip 0.6in}llr@{\hskip 0.6in}lr}

    \textbf{\textsc{paper \texttt{id}}} & \multicolumn{3}{c}{\textbf{\textsc{style}}} & \multicolumn{3}{c}{\textbf{\textsc{meaning}}} & \multicolumn{3}{c}{\textbf{\textsc{fluency}}} & \multicolumn{2}{c}{\textbf{\textsc{overall}}}  \\
    \addlinespace[0.2em]
    \toprule
    \addlinespace[0.5em]
    
    &  \texttt{metric} &  \texttt{arch.} &   & \texttt{metric} &  \texttt{arch.} &  & \texttt{metric} &  \texttt{arch.} &  & \texttt{metric} &  \\
    \addlinespace[0.3em]

     %2018 
     \textbf{\texttt{[1]}}   &   \textsc{reg}    & Linear reg. &  \xmark & \textsc{cls} & \cnn &  \cmark & \textsc{reg} & Linear reg. & \xmark & r-\bleu & \cmark\\
    %2019
    \textbf{\texttt{[2]}}   & \C  & \C & \C & \C  & \C & \C  & \C & \C & \C &  r-\bleu & $-$  \\  % 
    % 2019
    \textbf{\texttt{[3]}}   & \textsc{cls} &  \cnn &  $-$   & r-\bleu     & \C & $-$ & \C & \C & \C     & \textsc{gm(s,m)} & $-$ \\  
    % 2019
    \textbf{\texttt{[4]}}   & \textsc{cls} & \cnn & \xmark  &  r-\bleu    & \C &  \cmark & \C & \C & \C &  \textsc{gm(s,m)} &  \cmark  \\
    % 2019
    \textbf{\texttt{[5]}}   & \textsc{cls} &  \cnn &  \xmark &  r-\bleu & \C &  \cmark & \C & \C & \C  & \C & \C  \\
    % 2019
    \textbf{\texttt{[6]}}   & \textsc{cls} &  \lstm & $-$ & \textsc{cls}  & \bert &  $-$ & \C & \C & \C &   r-\bleu &   $-$ \\
    % 2019
    \textbf{\texttt{[7]}}   & \C    & \C    &  \C & \C & \C & \C & \C & \C & \C &  r-\bleu & $-$ \\
    % 2019
    \textbf{\texttt{[8]}}   & \textsc{cls} &  \cnn &  $-$ & \C & \C &  \C & \C & \C & \C & \C & \C   \\
    % 2019
   \textbf{\texttt{[9]}}    & \textsc{cls} & \lstm & $-$ & \textsc{emb-sim} & \C &  $-$ & \ppl & \textsc{lm (rnn)} &  $-$ & \textsc{f$1$(s,m)} & $-$ \\
    % 2020
   \textbf{\texttt{[10]}}   & \textsc{cls} &  \rob  & \xmark & \textsc{emb-sim} & \C & \cmark & \ppl &  \textsc{lm} (\rob)   &  $-$  & \textsc{j(s,m,f)} & \cmark\\  
    % 2020
   \textbf{\texttt{[11]}}   & \textsc{cls} & \cnn &  \xmark &  r-\bleu & \C &  \xmark & \C & \C & \C &  \textsc{f$1$(s,m)} & $-$ \\
    % 2020
    \textbf{\texttt{[12]}}  & \textsc{cls} &   \gru &  \xmark & \C &  \C & \C &  \textsc{cls} & Linear reg. & \xmark & r-\bleu & \xmark\\
    % 2020
    \textbf{\texttt{[13]}}  & \textsc{cls} &  \bert & \cmark &  r-\bleu & \C &  \cmark & \ppl &  \textsc{lm} (\kenlm) & \xmark & \textsc{gm(s,m,f)} &  $-$\\
    % 2020
    \textbf{\texttt{[14]}}  & \C   & \C    &  \C  & \C & \C  &  \C & \C &  \C & \C  & r-\bleu & $-$  \\
    % 2020 
    \textbf{\texttt{[15]}}  & \textsc{cls} &  \textsc{fastText} & \cmark &  r-\bleu & \C &  \cmark  & \ppl & \textsc{lm} (\gpt) & \cmark & \C & \C \\
    % 2020
    \textbf{\texttt{[16]}}  & \textsc{cls} &  \cnn     & $-$  &  r-\bleu & \C & $-$ & \ppl & \textsc{lm} (\lstm) & $-$ & \C & \C \\
    % 2020
    \textbf{\texttt{[17]}}  & \textsc{cls} &  \cnn & \cmark   & r-\bleu & \C &  \cmark & \C & \C & \C & \C &  \C \\
    % 2020
    \textbf{\texttt{[18]}}  & \textsc{cls} &  \cnn & \cmark &  r-\bleu & \C &  \cmark & \C & \C &  \C & \textsc{gm|hm(s,m)} &   \cmark  \\  %
    % 2020
   \textbf{\texttt{[19]}}   & \textsc{cls} & \gru   & $-$ & \textsc{cls} &  \bert &  $-$ & \C & \C & \C & r-\bleu &  \cmark \\
    % 2020
    \textbf{\texttt{[20]}}  & \textsc{cls} &  \rob &  $-$ &  r-\bleu & \C &  $-$ & \ppl & \textsc{lm} (\gpt) & $-$ & \textsc{gm|hm(s,m)} & $-$ \\
    % 2020
    \textbf{\texttt{[21]}}  & \textsc{cls} &  \cnn  & $-$     & r-\bleu & \C &  \cmark  & \ppl & \textsc{lm} (\gpt) & \cmark & \C & \C\\
    % 2021
    \textbf{\texttt{[22]}}  & \C   & \C  & \C  & \C & \C & \C & \C &  \C  & \C &  r-\bleu &  $-$  \\
    % 2021
    \textbf{\texttt{[23]}}  & \textsc{reg} &  \bert    & \xmark &  s-\bleu & \C &  \cmark  & \ppl & \textsc{lm} (\kenlm) & \xmark & r-\bleu &  \xmark  \\

    \end{tabular}}
    \caption{Details on automatic evaluation practices of prior work for \fost. For each dimension, \cmark \ \ and \xmark \ \ denote whether the best ranking system based on the automatic evaluation agrees or disagrees with the one pointed by the human evaluation; $-$ denotes that no human evaluation is conducted. \reg stands for regression, \textsc{cls} for classification, \textsc{sim} for similarity, \textsc{emb} for embedding-based, \textsc{lm} for language model, \textsc{gm} or \textsc{hm} for geometric or harmonic mean, \textsc{j} for corpus-level product, and \texttt{arch.} for architecture. The mappings between each paper \textsc{\texttt{id}} and its corresponding citation is included in Table~\ref{tab:status_mapping}.}
    \label{tab:status}
\end{table*}
\begin{table}[!t]
    \centering
    \scalebox{0.65}{
    \begin{tabular}{l@{\hskip 0.5in}l@{\hskip 0.8in}l}

    \textbf{\textsc{paper \texttt{id}}} & \textbf{\textsc{citation}} & \textbf{\textsc{language}} \\
    \addlinespace[0.2em]
    \toprule
    \addlinespace[0.5em]
    
     %2018 
    \textbf{\texttt{[1]}} &  \citet{rao-tetreault-2018-dear}         &   \english \\
    %2019
    \textbf{\texttt{[2]}} & \citet{kajiwara-2019-negative}          &   \english \\  % 
    % 2019
    \textbf{\texttt{[3]}} & \citet{li-etal-2019-domain}             & \english \\  
    % 2019
    \textbf{\texttt{[4]}} & \citet{ijcai2019-711}                   & \english \\
    % 2019
    \textbf{\texttt{[5]}} & \citet{shang-etal-2019-semi}            &  \english \\
    % 2019
    \textbf{\texttt{[6]}} & \citet{wang-etal-2019-harnessing}       &  \english \\
    % 2019
    \textbf{\texttt{[7]}} &  \citet{hybrid}                          &  \english \\
    % 2019
    \textbf{\texttt{[8]}} & \citet{korotkova}                       & \english, \textsc{et}, \textsc{lv}  \\
    % 2019
    \textbf{\texttt{[9]}} &  \citet{gong}                            & \english \\
    % 2020
    \textbf{\texttt{[10]}} & \citet{krishna-etal-2020-reformulating} & \english \\  
    % 2020
    \textbf{\texttt{[11]}} & \citet{abhilasha}                       &  \english \\
    % 2020
    \textbf{\texttt{[12]}} & \citet{Wu_Wang_Liu_2020}                &  \english \\
    % 2020
    \textbf{\texttt{[13]}} & \citet{ijcai2020-526}                   & \english \\
    % 2020
   \textbf{\texttt{[14]}} &  \citet{zhang-etal-2020-parallel}        &  \english \\
    % 2020
    \textbf{\texttt{[15]}} & \citet{goyal}                           & \english \\
    % 2020
    \textbf{\texttt{[16]}} & \citet{he2020a}                         &  \english \\
    % 2020
   \textbf{\texttt{[17]}} &  \citet{chen}                            & \english \\
    % 2020
    \textbf{\texttt{[18]}} & \citet{zhou}                            & \english \\  %
    % 2020
    \textbf{\texttt{[19]}} & \citet{wang-etal-2020-formality}        & \english \\
    % 2020
    \textbf{\texttt{[20]}} & \citet{Jingjing}                        & \english\\
    % 2020
    \textbf{\texttt{[21]}} & \citet{liu}                             & \english \\
    % 2021
    \textbf{\texttt{[22]}} & \citet{styleptb}                        & \english \\
    % 2021
    \textbf{\texttt{[23]}} & \citet{xformal}                         &  \portuguese, \french, \italian \\

    \end{tabular}}
    \caption{List of prior works on \fost along with languages (i.e., \texttt{iso} codes) addressed in each of them.}\vspace{-0.5cm}
    \label{tab:status_mapping}
\end{table}

\subsection{Limitations of Automatic Evaluation}

Recent work highlights the need for research to improve evaluation practices for \st along multiple directions. Not only does \st lack standardized evaluation practices \cite{yamshchikov2020styletransfer}, but commonly used methods have major drawbacks which hamper progress in this field.
\citet{pang-2019-towards} and \citet{Pang2019UnsupervisedEM} show that the most widely adopted automatic metric, \bleu, can be gamed. They observe that untransferred text achieves the highest \bleu score for the task of sentiment transfer, questioning complex models' ability to surpass this trivial baseline. \citet{mir-etal-2019-evaluating} discuss the inherent trade-off between \st evaluation aspects  and propose that models are evaluated at specific points of their trade-off plots.
\citet{tikhonov-etal-2019-style} argue that, despite their cost, human-written references
are needed for future experiments with style transfer. They also show  
that comparing models without reporting error margins can lead to incorrect conclusions as state-of-the-art models sometimes end up within error margins from one another. 

\subsection{Structured Review of \st Evaluation}\label{sec:lit_review}

We systematically review automatic evaluation practices in \st with formality as a case study.  We select \fost for this work since it is one of the most frequently studied styles~\cite{jin2021deep} and there is human annotated data including human references available for these evaluations \cite{rao-tetreault-2018-dear,xformal}.
Tables \ref{tab:status} and \ref{tab:status_mapping} summarize evaluation details for all \fost methods in papers from the \st survey by \citet{jin2021deep}.\footnote{The complete list is hosted at: \url{https://github.com/fuzhenxin/Style-Transfer-in-Text}}
Most works employ automatic evaluation for \textit{style}~($87\%$) and \textit{meaning} preservation ($83\%$). \textit{Fluency} is the least frequently evaluated dimension ($43\%$), while $74\%$
of papers employ automatic metrics to assess the \textit{overall} quality of system outputs that captures all desirable aspects. 

Across dimensions, papers also frequently rely on human evaluation ($55\%$, $58\%$, $60\%$, and $40\%$ for style, meaning, fluency, and overall). However, human judgments and automatic metrics do not always agree on the best-performing system. In $60\%$ of evaluations, the top-ranked system is the same according to human and automatic evaluation (marked as \cmark \ in Table~\ref{tab:status}), and their ranking disagrees in $40\%$ of evaluations (marked as \xmark \  in Table~\ref{tab:status}). When there is a disagreement, human evaluation is trusted more and viewed as the standard. This highlights the need for a systematic evaluation of automatic evaluation metrics.

Finally, almost all papers~($91\%$) consider \fost for English (\english), as summarized in Table~\ref{tab:status_mapping}. There are only two exceptions: 
\citet{korotkova} study \fost for Latvian (\textsc{lv}) and Estonian (\textsc{et}) in addition to \english,  
while \citet{xformal} 
study \fost for $3$ Romance languages: Brazilian Portuguese (\portuguese), French (\french
), and Italian (\italian). The former provides system output samples as a means of evaluation, and the latter employs human evaluations, highlighting the challenges of automatic evaluation in multilingual settings.

Next, we review the automatic metrics used for each dimension of evaluation in \fost papers. As we will see, a wide range of approaches is used. Yet, it remains unclear how they compare to each other, what their respective strengths and weaknesses are, and how they might generalize to languages other than English.

\subsection{Automatic Metrics for \fost}

\paragraph{Formality} Style transfer is often evaluated using model-based approaches. The most frequent method consists of training a binary classifier on human written formal vs. informal pairs. The classifier is later used to predict the percentage of generated outputs that match the 
desired attribute per evaluated system---the system with the highest percentage is considered the best
performing with respect to style. 
Across methods, the corpus used to train the classifier is the \textsc{gyafc} parallel-corpus~\cite{rao-tetreault-2018-dear} consisting of $105$K parallel informal-formal human-generated excerpts. This corpus is curated for \fost in \english, while similar resources
are not available for other languages. 
Different model architectures have been used by prior work (e.g., \cnn, \lstm, \gru, fine-tuning on pre-trained language models such as \rob and \bert; Table~\ref{tab:status}).
In most papers, the resulting classifier is evaluated on the test side of the \textsc{gyafc} corpus, reporting accuracy scores in the range of $80-90$\%.  Despite the high accuracy scores, the best ranking system under the classifier is very often in disagreement with human evaluations~(marked as \xmark \ \ under the third subcolumn of style of Table~\ref{tab:status}). A few works train regression-based models instead, using the training data of \citet{pavlick-tetreault-2016-empirical} that are human-annotated for formality on a $7$-point scale---while, again, this resource is only available for \english. 

\paragraph{Meaning Preservation} Evaluation of this dimension is performed using a wider spectrum of approaches, as presented in the third column of Table~\ref{tab:status}.  The most frequently used metric is reference-\bleu (r-\bleu), which is based on the $n$-gram precision of the system output compared to human rewrites of the desired formality. Other approaches include self-\bleu (s-\bleu), where the system output is compared to its input, measuring the semantic similarity between the system input and its output, or regression models (e.g., \cnn, \bert) trained on data annotated for similarity-based tasks, such as the Semantic Textual Similarity task (\sts)~\cite{agirre-etal-2016-semeval}. 

\paragraph{Fluency} Fluency is typically evaluated with model-based approaches (see fourth column of Table~\ref{tab:status}). 
Among those, the most frequent method is that of computing perplexity (\ppl) under a language model. 
The latter is either trained from scratch on the same corpus used to train the \fost models (i.e., \textsc{gyafc})
using different underlying architectures
(e.g., \kenlm, \lstm), or employ large pre-trained language models (e.g., \gpt). A few other works train models on \english data annotated for grammaticality \citep{heilman-etal-2014-predicting} or linguistic acceptability \citep{warstadt-etal-2019-neural} instead.

\paragraph{Overall} Systems' overall quality (see fifth column of Table~\ref{sec:status}) is mostly evaluated using r-\bleu or by combining independently computed metrics into a single score (e.g., geometric mean - \textsc{gm}(.), harmonic mean - \textsc{hm}(.), \textsc{f}$1$(.)). Moreover, $6$ out of $8$ approaches that rely on combined scores do not include fluency scores in their overall evaluation. 

\paragraph{English Focus}  Since most of the current work on \fost and \st is in \english, prior work relies heavily on \english resources for designing automatic evaluation methods. For instance, resources for training stylistic classifiers or regression models are not available for other languages.
For the same reason, it is unclear whether model-based approaches for measuring meaning preservation and fluency can be ported to multilingual settings. Furthermore, reference-based evaluations (e.g., r-\bleu) require human rewrites that are only available for \english, \portuguese, \italian, and \french. Finally, even though perplexity does not rely on annotated data, without standardizing the data language models are trained on, we cannot make meaningful cross-system comparisons.

\subsection{Summary}

Reviewing the literature shows the lack of standardized metrics
for \st evaluation, which hampers comparisons across papers, the lack of agreement between human judgments and automatic metrics, which hampers system development, and the lack of portability to languages other than English which severely limits the impact of the work. These issues motivate the controlled multilingual evaluation of evaluation metrics in our paper.

%%% Section 3: Methods
\section{Evaluating Evaluation Metrics}

We evaluate evaluation metrics (described in~\S\ref{sec:automatic_details}) for multilingual \fost, in four languages 
for which human evaluation judgments (described in~\S\ref{sec:human_details}) on \fost system outputs are available. 

\subsection{Human Judgments}\label{sec:human_details}

We use human judgments collected by prior work of ~\citet{rao-tetreault-2018-dear} for \english and \citet{xformal} for \portuguese, \french, and \italian. We include details on their annotation frameworks, the quality of human judges, and the evaluated systems below. 

\paragraph{Human Annotations} We briefly describe the annotation frameworks employed by~\citet{rao-tetreault-2018-dear} and \citet{xformal} to collect human judgments for each evaluation aspect: 
\begin{inparaenum}
    \item \textbf{formality} ratings are collected---for each system output---on a $7$-point discrete scale, ranging from $-3$ to $+3$, as per~\citet{DBLP:journals/corr/Lahiri15} 
    \textit{
    (Very informal, Informal, Somewhat Informal, Neutral, Somewhat Formal, Formal. Very Formal)};
    \item \textbf{meaning preservation} judgments adopt the Semantic Textual Similarity annotation scheme of~\citet{agirre-etal-2016-semeval}, where an
    informal input and its corresponding formal system output are rated on a scale from $1$ to $6$ based on their similarity
    \textit{
    (Completely dissimilar, Not equivalent but on same topic, Not equivalent but share some details, Roughly equivalent, Mostly equivalent, Completely equivalent)};
    \item \textbf{fluency} judgments are collected for each system output on a discrete scale of $1$ to $5$, as per~\citet{heilman-etal-2014-predicting} 
    \textit{
    (Other, Incomprehensible, Somewhat Comprehensible, Comprehensible, Perfect)};
    \item \textbf{overall} judgments are collected following a \textit{relative ranking} approach: all system outputs are ranked in order of their formality, taking into account both meaning preservation and fluency. 
\end{inparaenum}

\paragraph{Human Annotators} Both studies recruited workers 
from the Amazon Mechanical Turk platform after employing quality control methods to exclude poor quality workers (i.e., manual checks for \english, and qualification tests for \portuguese, \french, and \italian). For all human evaluations and languages \citet{xformal} report at least moderate inter-annotator agreement.

\paragraph{Evaluated Systems} The evaluated system outputs were sampled from $5$ \fost models for each language, spanning a range of simple baselines to neural architectures \cite{rao-tetreault-2018-dear, xformal}. We also include detailed descriptions of them in Appendix~\ref{sec:evaluated_systems}. For each evaluation dimension  $500$ outputs are evaluated for \english and $100$ outputs per system for \portuguese, \french, and \italian.

\subsection{Evaluation Metrics}\label{sec:automatic_details}

For the \fost evaluation aspects described below, we cover a broad spectrum of approaches that range from dedicated models for the tasks at hand to more lightweight methods relying on unsupervised approaches and automated metrics. 

\paragraph{Formality} We benchmark model-based approaches that fine-tune multilingual pre-trained language models (i.e., \textsc{xlm-r}, m\textsc{bert}), where the task of formality detection is modeled either as a \textbf{binary classification} task (i.e., formal vs. informal), or as a \textbf{regression} task that predicts different formality levels  on an ordinal scale. 

\paragraph{Meaning Preservation} We evaluate the \bleu score ~\cite{papineni-etal-2002-bleu} of the system output compared to the reference rewrite (r-\textsc{\bleu}) since it is the dominant metric in prior work. 
Prior reviews of meaning preservation metrics for paraphrase and sentiment \st tasks in \english \citep{yamshchikov2020styletransfer} cover $n$-gram metrics and embedding-based approaches.
We consider three additional metric classes to compare system outputs with inputs, as human annotators do:

\begin{enumerate}
    \item \textbf{n-gram based metrics} include: s-\textsc{bleu} (self-\bleu that compares system outputs with their inputs as opposed to references, i.e., r-\bleu), \textsc{meteor}~\cite{banerjee-lavie-2005-meteor} based on the harmonic mean of unigram precision and recall while accounting for synonym matches, and  
    chr\textsc{F}~\cite{popovic-2015-chrf} based on the character $n$-gram F-score;
    \item \textbf{embedding-based methods} fall under the category of unsupervised evaluation approaches that rely on either
    \textit{contextual} word representations extracted from pre-trained language models or 
    \textit{non-contextual pre-trained word embeddings} 
    (e.g., word2vec~\cite{w2v}; Glove~\cite{pennington-etal-2014-glove}). 
    For the former, we use \textsc{bert}-score~\cite{bertscore}
    which computes the similarity between each output token and each reference token based on \textsc{bert} contextual embeddings. For the latter, we experiment with two similarity metrics: the first is the cosine distance between the 
    sentence-level feature representations of the compared texts extracted via averaging their word embeddings; the second is the 
    \textit{Word Mover's Distance} (\textsc{wmd}) metric of~\citet{pmlr-v37-kusnerb15} that measures the dissimilarity between two texts as the minimum amount of distance that the embedded words of one text need to ``travel" to reach the word embeddings of the other;
    \item \textbf{semantic textual similarity (\textsc{sts}) models} 
    constitute supervised methods that we model via fine-tuning multilingual pre-trained language models (i.e., \textsc{xlm-r}, m\textsc{bert}) to predict a semantic similarity score for a pair of texts on an ordinal scale.

\end{enumerate}

\paragraph{Fluency} We experiment with \textbf{perplexity}~(\textsc{ppl}) and \textbf{likelihood}~(\textsc{ll}) scores based on probability scores of language models trained from scratch (e.g., Ken\textsc{lm}~\cite{heafield-2011-kenlm}), as well as \textbf{pseudo-likelihood scores}~(\textsc{pseudo-ll}) extracted from pre-trained masked language models similarly to \citet{salazar-etal-2020-masked}, by masking sentence tokens one by one. 

%%% Section 4: Conditions
\section{Experiment Settings}\label{sec:conditions}
\begin{table*}[!t]
    \centering
    \scalebox{0.83}{
    \begin{tabular}{lllllr}
    \rowcolor{gray!10}
    \textbf{\textsc{dimension}}& \textbf{\textsc{language code}} & \textbf{\textsc{dataset}} & \textbf{\textsc{lineage}}  & \textbf{\textsc{labels}} & \textbf{\textsc{size}} \\
    \addlinespace[0.2cm]
    %\midrule
    %\rowcolor{gray!10}
    %\multicolumn{5}{c}{\textit{formality}}\\    
    %\addlinespace[0.2cm]
    %
    \multirow{2}{*}{Formality} & \multirow{2}{*}{\english} & \textsc{gyafc}  & \citet{rao-tetreault-2018-dear} & informal vs. formal & $105$K\\
     & & Formality ratings & \citet{pavlick-tetreault-2016-empirical} & $[$ $-3$, $-2$, $-1$, $0$, $1$, $2$, $3$ $]$ & $5$K\\
     \addlinespace[0.2cm]
     %
     %\rowcolor{gray!10}
     %\multicolumn{5}{c}{\textit{meaning preservation}}\\ 
    \addlinespace[0.2cm]
    \multirow{1}{*}{Meaning} & \english & \textsc{sts} & \citet{cer-etal-2017-semeval}      & $[$ $1$, $2$, $3$, $4$, $5$ $]$ & $5$K \\
    % \addlinespace[0.2cm]
     
    % \rowcolor{gray!10}
    %\multicolumn{5}{c}{\textit{fluency}} \\
    \addlinespace[0.2cm]
    \multirow{1}{*}{Fluency} & \english, \textsc{br-pt}, \textsc{it}, \textsc{fr}  & OpenSubtitles &  \citet{lison-tiedemann-2016-opensubtitles2016}  & \textit{none} & $1$M \\
 
    \end{tabular}}
    \caption{Details on training data used for model-based metrics across the three \textsc{st} evaluation aspects.}
    \label{tab:training_data}
\end{table*}
\paragraph{Supervised Metrics} For all supervised model-based approaches,
we experiment with fine-tuning two multilingual pre-trained language models: 
\begin{inparaenum}
    \item multilingual \textsc{bert}, dubbed  \textbf{m\textsc{bert}}~\cite{devlin-etal-2019-bert}---a transformer-based model pre-trained
    with a masked language model objective on the concatenation of monolingual Wikipedia corpora from the $104$ languages with the largest Wikipedias. 
    \item \textbf{\textsc{xlm-r}}~\cite{conneau-etal-2020-unsupervised}---a transformer-based masked language model trained on $100$ languages using monolingual CommonCrawl data. 
\end{inparaenum} 
All models are based on the HuggingFace Transformers~\cite{wolf-etal-2020-transformers}\footnote{\url{https://github.com/huggingface/transformers}} library. We fine-tune with the Adam optimizer~\cite{KingmaB14}, a batch size of $32$, and a learning rate of $5\mathrm{e}{-5}$ for $3$ and $5$ epochs for classification and regression tasks, respectively. We perform a grid search on held-out validation sets over learning rate with values: $2\mathrm{e}{-3}$, $2\mathrm{e}{-4}$, $2\mathrm{e}{-5}$, and $5\mathrm{e}{-5}$ and over number of epochs with values: $3$, $5$, and $8$.

\paragraph{Cross-lingual Transfer} For supervised model-based methods that rely on the availability of human-annotated instances to train dedicated models for specific tasks, we experiment with three standard cross-lingual transfer approaches~(e.g., \citet{pmlr-v119-hu20b}):
\begin{inparaenum}
    \item \textsc{zero-shot} trains a single model on the \english training data and evaluates it on the original test data for each language;
    \item \textsc{translate-train} uses machine translation~(\textsc{mt}) to obtain training data in each language through translating the \english training set---and trains independent systems for each language;
    \item \textsc{translate-test} trains a single model on the \english training data and evaluates it on the test data that are translated into \english using \textsc{mt}.
\end{inparaenum}
%
%%%%%%%%%%%%%%%%%%%%%%%%%%%%%%%%%%%%%%%%%%%%%%%%%%%%%%%%%%%%%%%%%%%%%%%%%%%%%%
%
\paragraph{Unsupervised Metrics}  For meaning preservation metrics, we use the open-sourced implementations of: 
\citet{post-2018-call} for \bleu~\cite{papineni-etal-2002-bleu};
\citet{banerjee-lavie-2005-meteor} for \textsc{meteor};
\citet{popovic-2015-chrf} for chr\textsc{f}.\footnote{\url{https://github.com/mjpost/sacrebleu}}\textsuperscript{,}\footnote{\url{https://www.cs.cmu.edu/~alavie/METEOR/}}\textsuperscript{,}\footnote{\url{https://github.com/m-popovic/chrF}}
For \textsc{bert}-score we use the implementation of~\citet{bertscore};\footnote{\url{https://github.com/Tiiiger/bert_score}}
non-contextualized embeddings-based approaches are based on \texttt{fastText} pre-trained embeddings.\footnote{\url{https://fasttext.cc}}
For fluency metrics, we use the implementation
of~\citet{salazar-etal-2020-masked} for computing pseudo-likelihood.\footnote{\url{https://github.com/awslabs/mlm-scoring}}
\textsc{ppl} and \textsc{ll} scores are extracted from a
$5$-gram \kenlm model \citep{heafield-2011-kenlm}.\footnote{\url{https://github.com/kpu/kenlm}}

\paragraph{Training Data} Table~\ref{tab:training_data} presents 
statistics on the training data used for supervised and unsupervised models across the $3$ \textsc{st} evaluation aspects.  
For datasets that are only available for \english, we use 
the already available \textit{machine} translated resources for \textsc{sts} \footnote{\url{https://github.com/PhilipMay/stsb-multi-mt}} and formality datasets \cite{xformal}. The former employs the \texttt{DeepL} service (no information of translation quality is available) while the latter uses the \textsc{aws} translation service\footnote{\url{https://aws.amazon.com/translate}} (with reported \textsc{bleu} scores of $37.16$ (\portuguese), $33.79$ (\french), and $32.67$ (\italian)).\footnote{\bleu scores were computed on  $5{,}000$ randomly sampled data from OpenSubtitles.} The \kenlm models for all the languages are trained on $1$M randomly sampled sentences from the OpenSubtitles dataset \cite{lison-tiedemann-2016-opensubtitles2016}.

%%% Section 5: Results
\section{Experimental Results}

We analyze the results of comparing the outputs from the several automatic metrics to their human-generated counterparts for formality style transfer (\S\ref{sec:formality_results}), meaning preservation (\S\ref{sec:meaning_results}), fluency (\S\ref{sec:fluency_results}) via conducting segment-level analysis---and then, turn into analyzing system-level rankings to evaluation overall task success (\S\ref{sec:overall_results}).

%%%%%% FORMALITY %%%%%%%%%%%%%%%
\begin{table*}[!t]
    \centering
    \scalebox{0.85}{
    \begin{tabular}{lllllllllrrrr}
    
    \rowcolor{gray!10}
    \textbf{\textsc{method}} & \multicolumn{4}{c}{\textbf{m\textsc{bert}}} & \multicolumn{4}{c}{\textbf{\textsc{xlm-r}}} & \multicolumn{4}{c}{{$\mathbf{\delta}$\textbf{(\textsc{xlm-r}, m\textsc{bert})}}} \\
    
    \cmidrule(l){1-13}
                                  & 
                                  \multicolumn{1}{l}{\english} &
                                  \multicolumn{1}{l}{\portuguese} & 
                                  \multicolumn{1}{l}{\french} & 
                                  \multicolumn{1}{l}{\italian} & 
                                  \multicolumn{1}{l}{\english} & 
                                  \multicolumn{1}{l}{\portuguese} & 
                                  \multicolumn{1}{l}{\french} & 
                                  \multicolumn{1}{l}{\italian}&
                                  \multicolumn{1}{l}{\english} & 
                                  \multicolumn{1}{l}{\portuguese} & 
                                  \multicolumn{1}{l}{\french} & 
                                  \multicolumn{1}{l}{\italian}\\
    \cmidrule(l){2-13}
    
    \textsc{zero-shot}       &  $90$  &  $87$ & $84$ & $88$ & $90$   & $90$  & $86$   & $89$    & $0$ & {\color{darkgreen}$+3$} & 
    {\color{darkgreen}$+2$} & {\color{darkgreen}$+1$}\\
    \textsc{translate-train} &        &  $85$ {\small\color{rr}{($\mathbf{-2}$)}} & $82$  {\small\color{rr}{($\mathbf{-2}$)}} & $83$
    {\small\color{rr}{($\mathbf{-5}$)}} & & $87$ {\small\color{rr}{($\mathbf{-3}$)}} & $84$ {\small\color{rr}{($\mathbf{-2}$)}}
    & $84$ {\small\color{rr}{($\mathbf{-5}$)}} & & {\color{darkgreen}$+2$} & {\color{darkgreen}$+2$} & {\color{darkgreen}$+1$}  \\
    
    \textsc{translate-test}  &        &  $82$ {\small\color{rr}{($\mathbf{-5}$)}} & $73$ {\small\color{rr}{($\mathbf{-11}$)}} &
    $79$ {\small\color{rr}{($\mathbf{-11}$)}} & & $82$ {\small\color{rr}{($\mathbf{-8}$)}} & $74$ {\small\color{rr}{($\mathbf{-12}$)}} & $80$ {\small\color{rr}{($\mathbf{-9}$)}} & & $0$ & {\color{darkgreen}$+2$} & {\color{darkgreen}$+1$}
    \\
    \end{tabular}}
    \caption{F1 scores of binary formality classifiers under different cross-lingual transfer settings. Numbers in parentheses indicate performance drops over \textsc{zero-shot}. \textsc{zero-shot} yields the highest scores across languages and pre-trained language models. \textsc{xlm-r} yields improvements over m\textsc{bert} across most setting ($\delta$(\textsc{xlm-r}, m\textsc{bert})).}
    \label{tab:binary_classifiers_results}
\end{table*}
\begin{figure*}[!t]
    \centering
    \subfigure[\textsc{en}]{\includegraphics[width=0.24\textwidth]{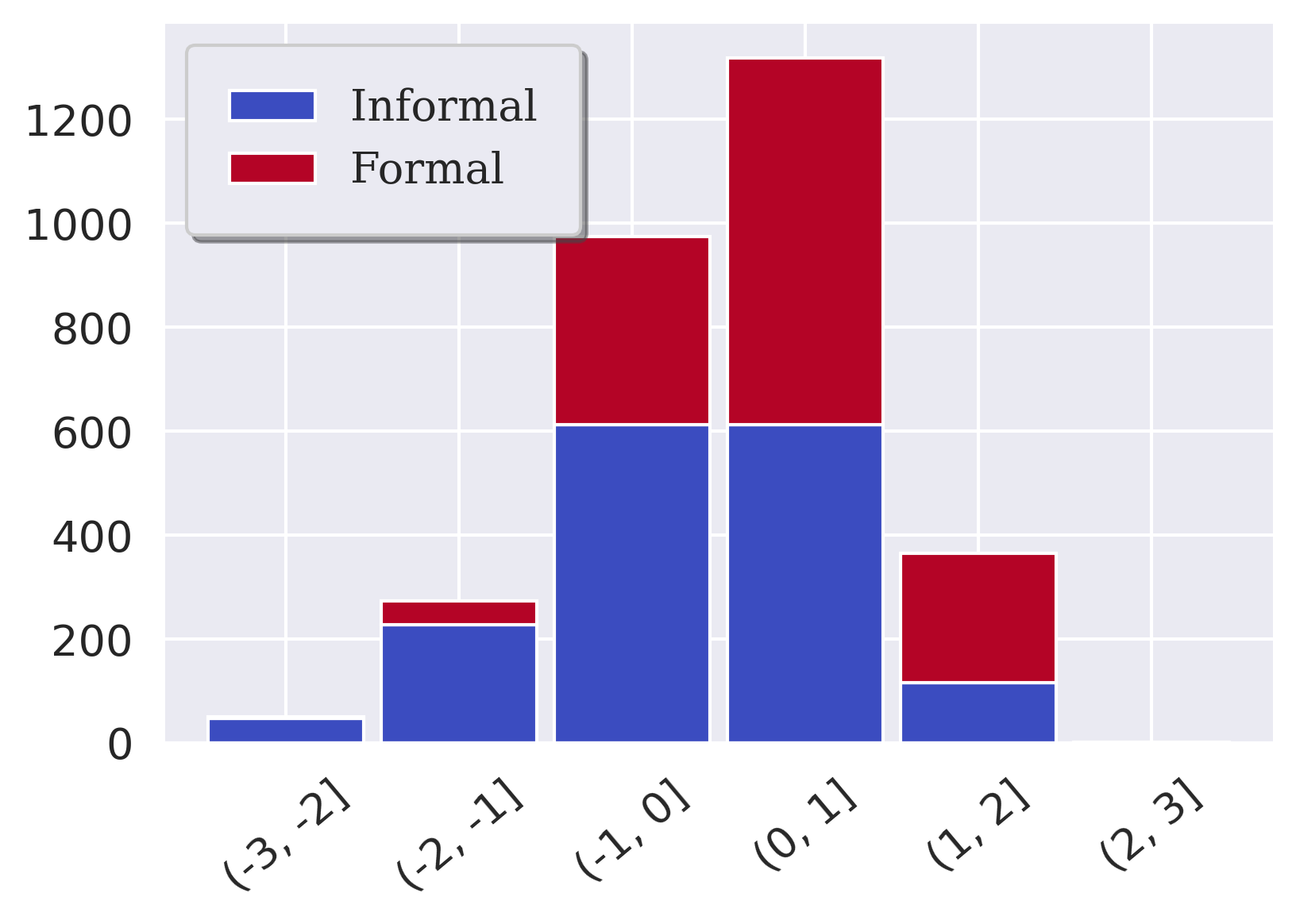}} 
    \subfigure[\textsc{br-pt}]{\includegraphics[width=0.24\textwidth]{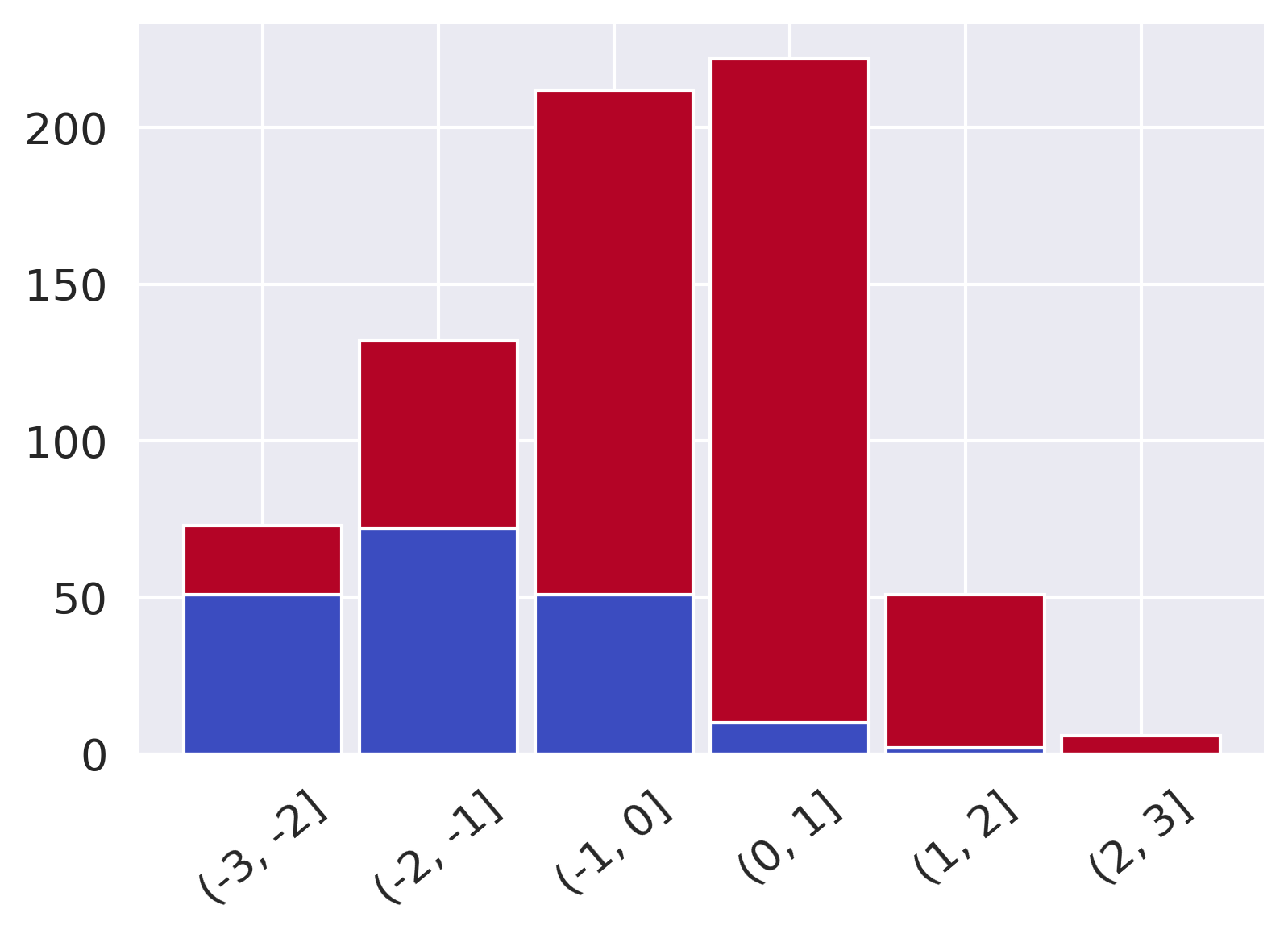}} 
    \subfigure[\textsc{fr}]{\includegraphics[width=0.24\textwidth]{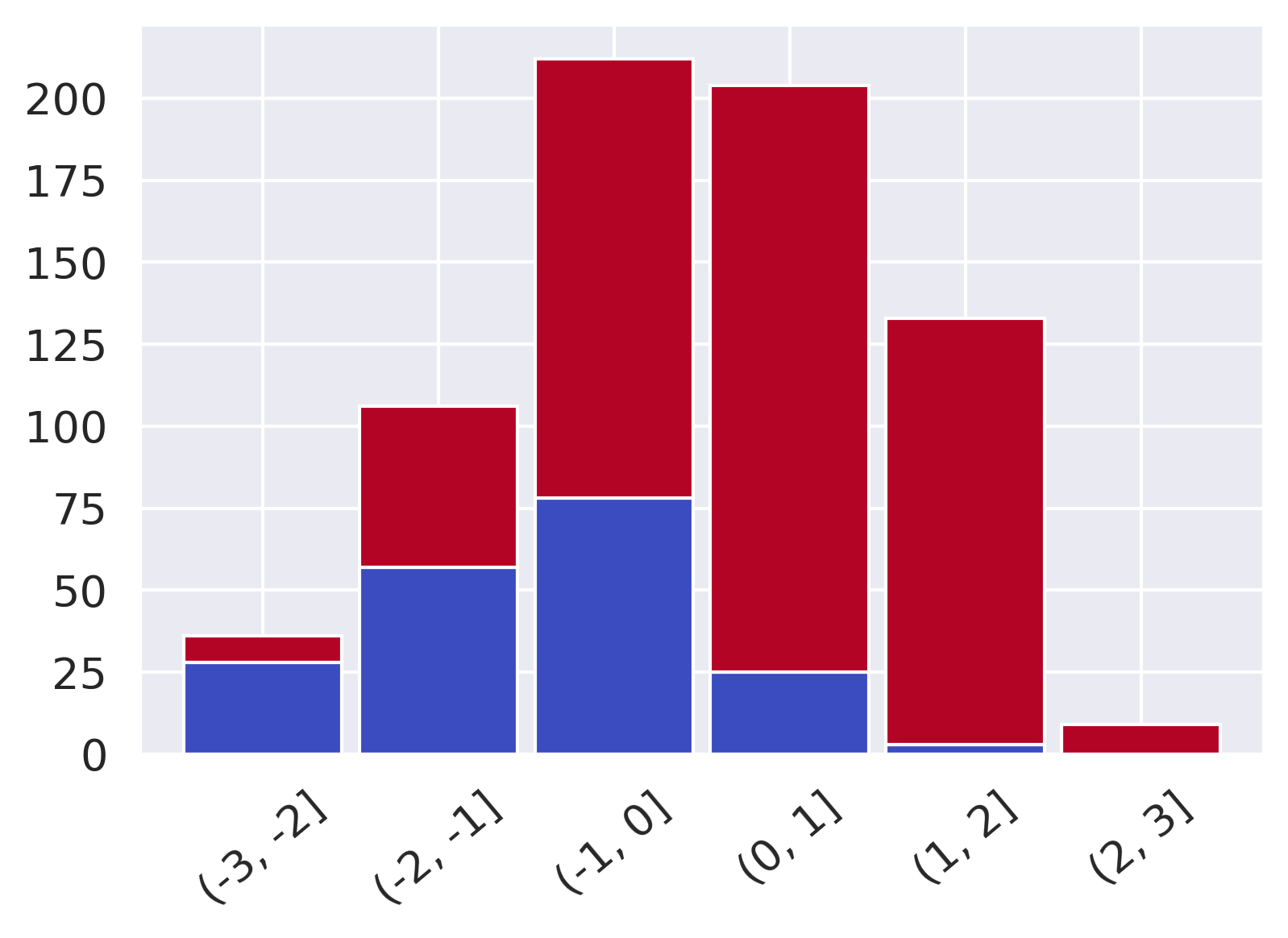}} 
    \subfigure[\textsc{it}]{\includegraphics[width=0.24\textwidth]{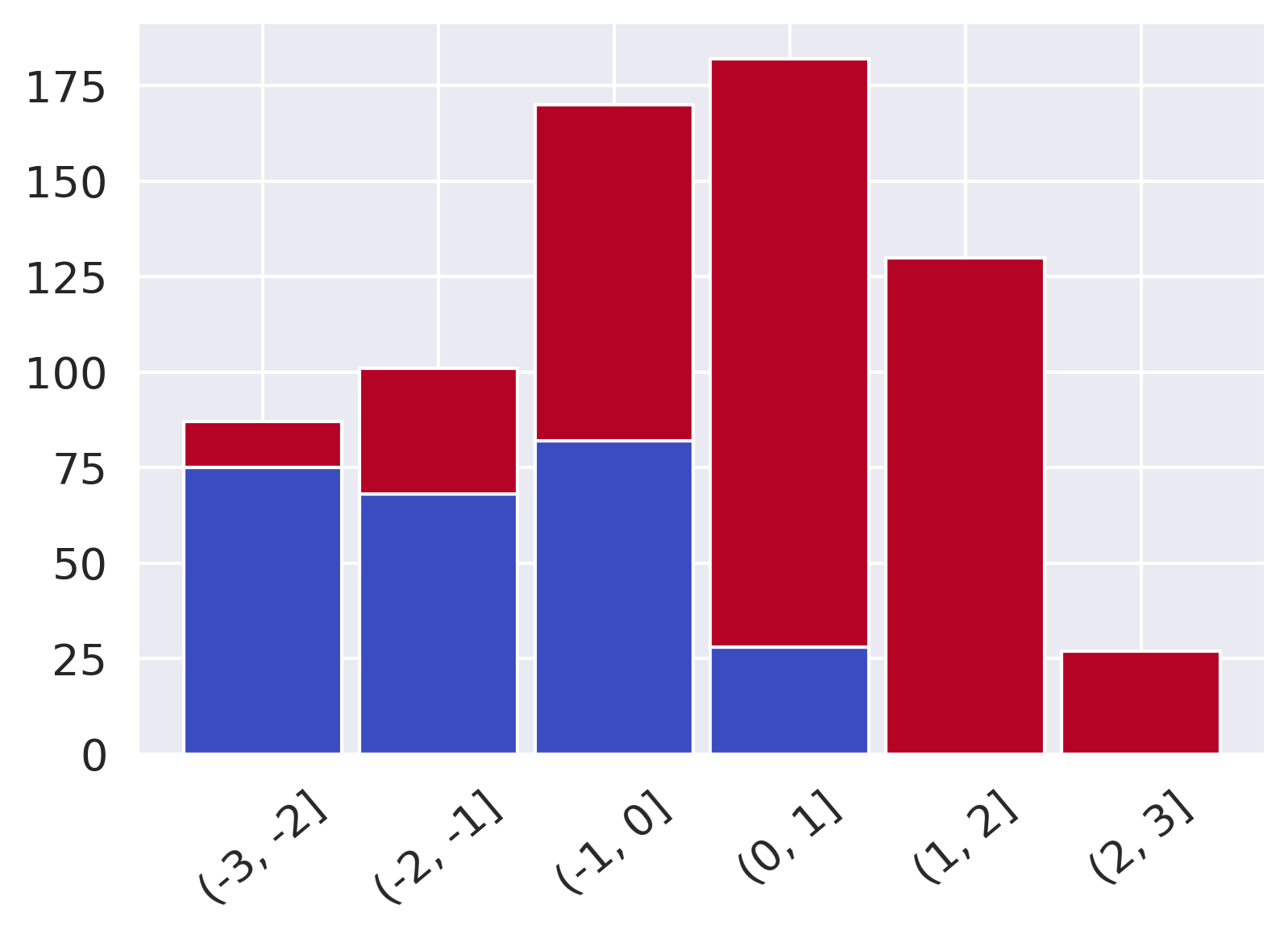}} 
    \caption{Number of formal vs. informal predictions of binary classifiers for each formality bin. Binary classifiers are biased towards outputing formal predictions for \textsc{br-pt, fr}, and \textsc{it}, while their performance degrades when moving closer to more neutral formality levels.}
    \label{fig:counts_regression}\vspace{0.05cm}
\end{figure*}
\begin{table*}[!t]
    \centering
    \scalebox{0.85}{
    \begin{tabular}{lllllllllrrrr}
    
    \rowcolor{gray!10}
    \textbf{\textsc{method}} & \multicolumn{4}{c}{\textbf{m\textsc{bert}}} & \multicolumn{4}{c}{\textbf{\textsc{xlm-r}}} & \multicolumn{4}{c}{{$\mathbf{\delta}$\textbf{(\textsc{xlm-r}, m\textsc{bert})}}} \\
    \cmidrule(l){1-13}

                                  & 
                                  \multicolumn{1}{l}{\english} &
                                  \multicolumn{1}{l}{\portuguese} & 
                                  \multicolumn{1}{l}{\french} & 
                                  \multicolumn{1}{l}{\italian} & 
                                  \multicolumn{1}{l}{\english} & 
                                  \multicolumn{1}{l}{\portuguese} & 
                                  \multicolumn{1}{l}{\french} & 
                                  \multicolumn{1}{l}{\italian}&
                                  \multicolumn{1}{l}{\english} & 
                                  \multicolumn{1}{l}{\portuguese} & 
                                  \multicolumn{1}{l}{\french} & 
                                  \multicolumn{1}{l}{\italian}\\
                                  
    \cmidrule(l){2-13}

    \textsc{zero-shot}       &  $66$  &  $64$ & $51$ & $51$ & $67$   & $72$  & $59$   & $61$    & {\color{darkgreen}$+1$} & {\color{darkgreen}$+8$} & 
    {\color{darkgreen}$+8$} & {\color{darkgreen}$+10$}\\
    \textsc{translate-train} &        &  $61$ {\small\color{rr}{($\mathbf{-3}$)}} & $48$  {\small\color{rr}{($\mathbf{-3}$)}} & $49$
    {\small\color{rr}{($\mathbf{-2}$)}} & & $63$ {\small\color{rr}{($\mathbf{-9}$)}} & $52$ {\small\color{rr}{($\mathbf{-7}$)}}
    & $55$ {\small\color{rr}{($\mathbf{-6}$)}} & & {\color{darkgreen}$+2$} & {\color{darkgreen}$+4$} & {\color{darkgreen}$+6$}  \\
    
    \textsc{translate-test}  &        &  $60$ {\small\color{rr}{($\mathbf{-4}$)}} & $39$ {\small\color{rr}{($\mathbf{-12}$)}} &
    $45$ {\small\color{rr}{($\mathbf{-10}$)}} & & $61$ {\small\color{rr}{($\mathbf{-11}$)}} & $37$ {\small\color{rr}{($\mathbf{-15}$)}} & $51$ {\small\color{rr}{($\mathbf{-10}$)}} & & {\color{darkgreen}$+1$} & {\color{rr}$-2$} & {\color{darkgreen}$+6$}
    \\
    \end{tabular}}
    \caption{Spearman's $\rho$ correlation (\%) of formality regression models. %under different cross-lingual transfer settings. 
    Numbers in parentheses indicate performance drops over \textsc{zero-shot}. \textsc{zero-shot} yields the highest scores across languages and pre-trained language models. \textsc{xlm-r} yields improvements over m\textsc{bert} across most settings ($\delta$(\textsc{xlm-r}, m\textsc{bert})).}
    \label{tab:regression_results}
\end{table*}

\subsection{Formality Transfer Metrics}
\label{sec:formality_results} 

The field is divided on the best way to evaluate the style dimension \---\ formality in our case. 
Practitioners use either a binary approach (is the new sentence formal or informal?) or a regression approach (how formal is the new sentence?).  We discuss the first approach and its limitations in \S~\ref{sec:binary_classifiers}, before moving to regression in \S~\ref{sec:regression_models}.

\subsubsection{Evaluating Binary Classifiers}\label{sec:binary_classifiers} 

As discussed in~\S\ref{sec:status}, the vast majority of \fost works evaluate style transfer based on the accuracy of a binary classifier trained to predict whether human-written segments are formal or informal. 
Yet, as Table~\ref{tab:status} indicates, this approach fails to identify the best system in this dimension 59\% of the time.
To better understand this issue, we evaluate these classifiers on human-written texts versus \textsc{st} system outputs.

\paragraph{Human Written Texts} Table~\ref{tab:binary_classifiers_results} presents F$1$ scores when testing the binary formality classifiers on the task they are trained on: predicting whether human-written sentences from \gyafc and \xformal are formal or informal. First, the last column (i.e., $\delta$(\textsc{xlm-r}, m\textsc{bert})) shows that \textsc{xlm-r} is a better model than m\textsc{bert} for this task, across languages, with the largest improvements in the \textsc{zero-shot} setting where \textsc{xlm-r} beats m\textsc{bert} by $+3$, $+2$, $+1$ for \portuguese, \french, and \italian respectively.

Second, \textsc{zero-shot} is surprisingly the best strategy to port \english models to other languages.  \textsc{translate-train} and \textsc{translate-test} hurt F1 by $3$ and $9$ points on average compared to \textsc{zero-shot}, despite exploiting more resources in the form of machine translation systems and their training data. However, transfer accuracy is likely affected by regular translation errors (as suggested by larger F1 drops for languages with lower \textsc{mt} \bleu scores) and by formality-specific errors. Machine translation has been found to produce outputs that are more formal than its inputs \citep{xformal}, which yields noisy training signals for \textsc{translate-train} and alters the formality of test samples for \textsc{translate-test}.  

\paragraph{System Outputs} We now evaluate the best performing binary classifier (i.e., \textsc{xlm-r} in \textsc{zero-shot} setting) on real system outputs---a setup in line with automatic evaluation frameworks. 
Figure~\ref{fig:counts_regression} presents a breakdown of the number of formal vs. informal predictions of the classifiers binned by human-rated formality levels. Across languages, the performance of the classifier deteriorates as we move away from extreme formality ratings (i.e., very informal ($-3$) and very formal ($+3$)). This lack of sensitivity to different formality levels is problematic since system outputs across languages are concentrated around neutral formality values. In addition, when testing on \portuguese, \french, and \italian (\textsc{zero-shot} settings), the classifier is more biased towards the formal class, which leads one to question its ability to correctly evaluate more formal outputs in multilingual settings. 
Taken together, these results suggest that validating the classifiers against human rewrites rather than system outputs is unrealistic and potentially misleading.

\subsubsection{Regression Models}\label{sec:regression_models} 
Table~\ref{tab:regression_results} presents Spearman’s $\rho$ correlation of
regression models' predictions with human judgments. Again, \textsc{xlm-r} with \textsc{zero-shot} transfer yields the highest correlation across languages. More specifically, the trends across different transfer approaches and different pre-trained language models are similar to the ones observed on evaluation of binary classifiers: 
\textsc{xlm-r} outperforms m\textsc{bert} for almost all settings, while \textsc{zero-shot} is the most successful
transfer approach, followed by \textsc{translate-train}, with \textsc{translate-test} yielding the lowest correlations across languages. Interestingly, regression models highlight the differences between the generalization abilities of \textsc{xlm-r} and m\textsc{bert} more clearly than the previous analysis on binary predictions: \textsc{zero-shot} transfer on \textsc{xlm-r} yields $8\%$, $8\%$, and $10\%$ higher correlations than m\textsc{bert} for \portuguese, \french, and \italian---while both models yield similar correlations for \english.

\subsection{Meaning Preservation Metrics}
\label{sec:meaning_results} 
%%%%%% MEANING (REFERENCE-FREE). %%%%%%%%%%%%%%%%
\begin{table}[!t]
    \centering
    \scalebox{0.91}{
    \begin{tabular}{llcccc}
    \rowcolor{gray!10}
    & \textbf{\textsc{metric}}     & \textbf{\textsc{en}} & \textbf{\textsc{it}} & \textbf{\textsc{br-pt}} & \textbf{\textsc{fr}} \\
    \addlinespace[0.2cm]
    
    & r-\textsc{bleu}   & $0.31$ & $0.00$ & $0.05$ & $0.12$ \\
    
    & s-\textsc{bleu}   & $0.48$ & $0.59$ & $0.67$ & $0.66$ \\
    & \textsc{meteor} & $0.54$ & $0.58$ & $0.64$ & $0.60$ \\
    & chr\textsc{f} & $\mathbf{0.59}$ & $\mathbf{0.67}$ & $\mathbf{0.70}$ & $\mathbf{0.75}$ \\
        
    \addlinespace[0.3cm]

    & \textsc{wmd} & $0.51$ & $0.52$ & $0.57$ & $0.55$ \\
            \rot{\rlap{~\textit{unsupervised}}} 
    & Cosine &  $0.49$ & $0.50$ & $0.54$ & $0.52$ \\
    & \bert-score & $0.59$ & $0.62$ & $0.65$ & $0.68$ \\
        
    \addlinespace[0.3cm]
    \cmidrule(l){2-6}
    &  \multicolumn{5}{c}{\textsc{translate-train (sts)}}\\
    \cmidrule(l){2-6}
    & m-\bert  & $0.53$ & $0.58$ & $0.51$ & $0.63$ \\
    & \textsc{xlm-r}& $0.55$ & $0.64$ & $0.57$ & $0.60$ \\

    \addlinespace[0.1cm]

    \cmidrule(l){2-6}
    & \multicolumn{5}{c}{\textsc{zero-shot (sts)}}\\
    \cmidrule(l){2-6}
                \rot{\rlap{~\textit{supervised}}} 
    & m-\bert  & - & $0.61$ & $0.60$ & $0.66$ \\
    & \textsc{xlm-r}& - & $\mathbf{0.67}$ & $0.68$ & $0.65$ \\
    \end{tabular}}
    \caption{Spearman's $\rho$ correlation of meaning preservation metrics with human judgments.}
    \label{tab:meaning_free}\vspace{-0.3cm}
\end{table}
Table~\ref{tab:meaning_free} presents Spearman's $\rho$ correlation of meaning preservation metrics with human judgments.  chr\textsc{F}
consistently yields the highest correlations across languages---this result is in line with prior observations on evaluating meaning preservation metrics for \english \st tasks~\cite{yamshchikov2020styletransfer} and is now confirmed in a multilingual setting. 
This trend might be explained by chr\textsc{f}'s ability to match spelling errors within words via character $n$-grams. 
\textsc{xlm-r} trained on \sts with zero-shot transfer is a close second to chr\textsc{f}, consistent with this model's top-ranking behavior as a formality transfer metric. 
However, chr\textsc{f} outperforms the remaining more complex and expensive metrics, including \textsc{bert}-score and m\textsc{bert} models.   
In contrast to \citet{yamshchikov2020styletransfer}, 
embedding-based methods (i.e., cosine, \textsc{wmd}) show no advantage over $n$-gram metrics, perhaps due to differences in word embedding quality across languages.  Finally, it should be noted that r\textsc{-bleu} is the worst performing metric across languages, and its correlation with human scores is particularly poor for languages other than English. This is remarkable because it has been used in $75\%$ of automatic evaluations for \fost meaning preservation evaluation (as seen in Table~\ref{tab:status}). We, therefore, recommend discontinuing its use.

%%%%%% FLUENCY %%%%%%%%%%%%%%%
\subsection{Fluency Metrics}
\label{sec:fluency_results} 
Table~\ref{tab:fluency_correlations_with_human} presents Spearman's $\rho$ correlation of various fluency metrics with human judgments.
Pseudo-likelihood (\textsc{pseudo-ll}) scores obtained from \textsc{xlm-r} correlate with human fluency ratings best across languages. Their correlations are strong across languages, while other methods only yield weak (i.e., Ken\textsc{lm}, m\textsc{bert}) to moderate correlations (i.e, Ken\textsc{lm-ppl}) for \italian. We, therefore, recommend evaluating fluency using Pseudo-likelihood scores derived from \textsc{xlm-r} to help standardize fluency evaluation across languages. 

\begin{table}[!t]
    \centering
    \scalebox{0.91}{
    \begin{tabular}{lcccc}
    
    \rowcolor{gray!10}
    \multicolumn{1}{l}{\textbf{\textsc{method}}} & \textbf{\textsc{en}} & \textbf{\textsc{it}} & 
    \textbf{\textsc{pt}} & \textbf{\textsc{fr}} \\
    \addlinespace[0.5em]
    
    \kenlm (\textsc{ll}) & $0.33$ & $0.27$ &	$0.43$ & 	$0.39$\\
    \kenlm (\textsc{ppl}) & $0.40$ & $0.35$ &	$0.45$ & 	$0.41$\\ 
    
    \addlinespace[0.1cm]
    
    m\textsc{bert} (\textsc{pseudo-ll}) & $0.42$ & 	$0.28$	& $0.43$ &	$0.41$ \\
    \textsc{xlm-r} (\textsc{pseudo-ll}) & $\mathbf{0.50}$	& $\mathbf{0.46}$& 	$\mathbf{0.55}$ &	$\mathbf{0.61}$\\

    \end{tabular}}
    \caption{Spearman's $\rho$ correlation of fluency metrics with human judgments.}
    \label{tab:fluency_correlations_with_human}\vspace{-0.3cm}
\end{table}

%%%%%% Cross-methods %%%%%%%%%%%%%%%

\subsection{System-level Rankings}
\label{sec:overall_results} 

Finally, we turn to predict the overall ranking of systems by focusing on how many correct pairwise system comparisons each metric gets correct.  For each language, there are $5$ systems, which means there are $10$ pairwise comparisons, for a total of $40$ given the $4$ languages.  We analyze corpus-level r\textsc{-bleu}, commonly used for this dimension, along with leading metrics from the other dimensions: \textsc{xlm-r} formality regression models, chr\textsc{f} and \textsc{xlm-r} pseudo-likelihood.  r\textsc{-bleu} gets $30$ out of $40$ comparisons correct while the other metrics get $25$, $22$, and $19$ respectively. This indicates that r\textsc{-bleu} correlates with human judgments better at the corpus-level than at the sentence-level, as in  machine translation evaluation \citep{mathur-etal-2020-results}.  We caution that these results are not definitive but rather suggestive of the best performing metric, given the
ideal evaluation would be a larger number of systems with which to perform a rank correlation. The complete analysis for each language is in Appendix~\ref{sec:pairwise_results}.

%%% Section 6: Conclusion
\section{Conclusions}

Automatic (and human) evaluation processes are well-known problems for the field of Natural Language Generation \cite{howcroft-etal-2020-twenty, clinciu-etal-2021-study} and the burgeoning subfield of \st is not immune.  \st, in particular, has suffered from a lack of standardization of automatic metrics, a lack of agreement between human judgments and automatics metrics, as well as a blindspot to developing metrics for languages other than English.  
We address these issues by conducting the first controlled multilingual evaluation for leading \st metrics with a focus on formality, covering metrics for $3$ evaluation dimensions and overall ranking for $4$ languages. Given our findings, we recommend the formality style transfer community adopt the following best practices:

\paragraph{1. Formality} \textsc{xlm-r} formality regression models in the \textsc{zero-shot} cross-lingual transfer setting yields the clear best metrics across all four languages as it correlates very well with human judgments.
However, the commonly used binary classifiers do not generalize across languages (due to misleadingly over-predicting formal labels). We propose that the field use regression models instead since they are designed to capture a wide spectrum of formality rates.

\paragraph{2. Meaning Preservation} We recommend using chr\textsc{f} as it exhibits strong correlations with human judgments for all four languages.  
We caution against using \bleu for this dimension, despite its overwhelming use in prior work as both its reference and self variants do not correlate as strongly as other more recent metrics.

\paragraph{3. Fluency} \textsc{xlm-r} is again the best metric (in particular for French). However, it does not correlate well with human judgments as compared to the other two dimensions.  

\paragraph{4. System-level Ranking} chr\textsc{f} and  \textsc{xlm-r} are the best metrics using a pairwise comparison evaluation.  However, an ideal evaluation would be to have a large number of systems with which to draw reliable correlations.

\paragraph{5. Cross-lingual Transfer} Our results support using
\textsc{zero-shot} transfer 
instead of machine translation to port metrics from English to other languages for formality transfer tasks.\\

We view this work as a strong point of departure for future investigations of \st evaluation. Our work first calls for 
 exploring how these evaluation metrics generalize to other \textit{styles} and \textit{languages}.  Across the different ways of defining style evaluation (either automatic or human), prior work has mostly focused on the three main dimensions covered in our study. As a result, although our meta-evaluation on \textsc{st} metrics focuses on formality as a case study, it can inform the evaluation of other style definitions (e.g., politeness, sentiment, gender, etc.). However, more empirical evidence is needed to test the applicability of the best performing metrics for evaluating style transfer beyond formality. 
Our work suggests that the top metrics based on \textsc{xlm-r} and chr\textsc{f} are robust across $4$ Romance languages; yet, our conclusions and recommendations are currently limited to \textit{this} set of languages. We hope that future work in multilingual style transfer will allow for testing their generalization to a broader spectrum of languages and style definitions.  
Furthermore, our study highlights that more research is needed on automatically ranking systems. For example, one could build a metric that combines metrics' outputs for the three dimensions, or one could develop a singular metric.
In line with \citet{briakou-etal-2021-review}, our study also calls for releasing more human evaluations and more system outputs to enable robust evaluation.
Finally, there is still room for improvement in assessing how fluent a rewrite is. 
Our study provides a framework to address these questions systematically and calls for \st papers to standardize and release data to support larger-scale evaluations.

\section*{Acknowledgements}

We thank 
Sudha Rao for providing references and materials of the \textsc{gyafc} dataset,
Jordan Boyd-Graber, Pedro Rodriguez, the \textsc{clip} lab at \textsc{umd},  and the
\textsc{emnlp} reviewers for their helpful and constructive comments.

\bibliography{anthology,custom}
\bibliographystyle{acl_natbib}

\clearpage
\appendix

\section{System-level Analysis}\label{sec:pairwise_results}

Table~\ref{tab:system_pairise} presents the number of correct system-level pair-wise comparisons of automatic metrics based on human judgments. For \textsc{sts}, chr\textsc{f}, \textsc{f.reg*}, \textsc{f.class*}, and \textsc{pseudo-lkl*}, system-level scores are extracted via averaging sentence-level scores. For s-\bleu and r-\bleu the system scores are extracted at the corpus-level. The total number of pairwise comparisons for each language is $10$ (given access to $5$ systems). Among the meaning preservation metrics (i.e., \textsc{sts}, s-\bleu, and chr\textsc{f}), chr\textsc{f} yields the highest number of correct comparisons (i.e., $37$ out of $40$ for all languages). The formality regression models (i.e., \textsc{f.reg*}) result in correct rankings more frequently than the formality classifiers (i.e., \textsc{f.class*}) yielding $35$ out of $40$ correct comparisons. Reference-\bleu (i.e., r-\bleu) is compared with overall ranking judemnts. It  ranks $8$ out of $10$ systems correctly for \english, \french, and \portuguese and only $6$ for \italian. Finally, perplexity (i.e., \textsc{ppl}) results in the fewest correct rankings at system-level (i.e., $22$ out of $40$), despite correlating well with human judgments at the segment-level.

\begin{figure}[!t]
    \centering
    \includegraphics[width=0.4\textwidth]{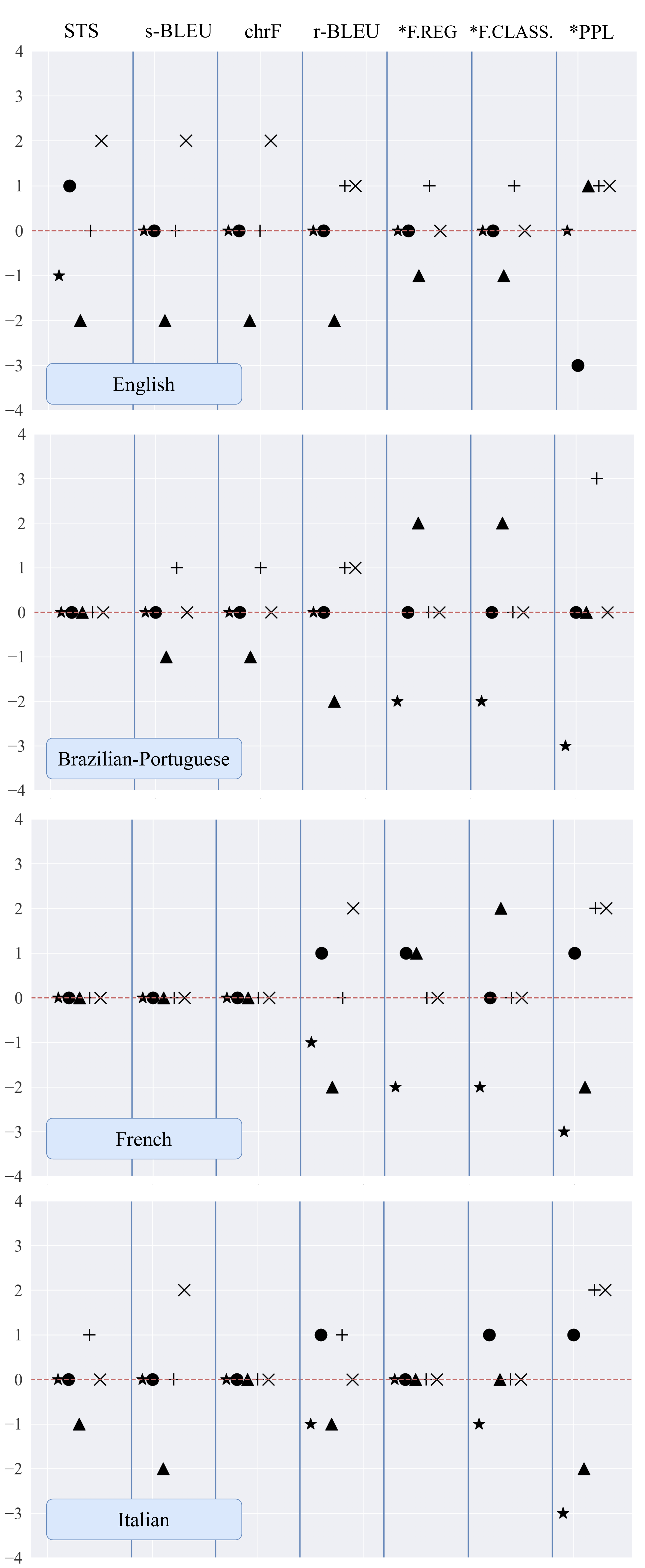}
    \caption{Difference in relative ranking between human judgments and automatic metrics across systems (i.e, represented by different markers) for different evaluation dimensions. 
    \sts, s-\bleu and ch\textsc{rf} are compared with meaning rankings, r-\bleu (reference-\bleu) with overall, \textsc{xlm-r} classifiers (*\textsc{f.class}) and regression (*\textsc{f.ref}) models with formality, and \textsc{xlm-r} pseudo-perplexity (*\textsc{ppl}) with fluency.
    }
    \label{fig:system_ranking_per_dimenstion}
\end{figure}

Additionally, in Figure~\ref{fig:system_ranking_per_dimenstion} we visualize the differences between relative rankings induced by human judgments and the best segment-level correlated metrics for each dimension, averaged per system. 

\begin{table*}[!t]
    \centering
    \scalebox{0.8}{
    \begin{tabular}{lrrrrrrrrrrr}
    \rowcolor{gray!10}
                & \multicolumn{4}{c}{\textsc{meaning}} & \multicolumn{2}{c}{\textsc{formality}} & \textsc{fluency} & \multicolumn{4}{c}{\textsc{overall}}\\
    Language     & \textsc{sts} & s-\textsc{bleu} & r-\textsc{bleu} & chr\textsc{f} & \textsc{f.reg*} & \textsc{f.class*} & \textsc{ppl}  &    r-\textsc{bleu} &  \textsc{f.reg*} &  chr\textsc{f} & pseudo-\textsc{ll} \\
    \textsc{en}  & 6  & 7  & 8 & 7  & 9  & 9 & 7 & 8 & 6 & 6 & 3\\
    \portuguese  & 10 & 9  & 7 & 9  & 7  & 7 & 5 & 8 & 8 & 4 & 5\\
    \french      & 9  & 7  & 10 & 10 & 10 & 9 & 5 & 8 & 6 & 3 & 7\\
    \italian     & 10 & 10 & 8 &10 & 9  & 8 & 5 & 6 & 5 & 9 & 4\\
    \textsc{all} & 35 & 33 & 35 & 36 & 35 & 33 & 22 & 30 & 25 & 22 & 19\\ 
    \end{tabular}}
    \caption{Number of correct system level pair-wise comparisons between $5$ systems for each language.}
    \label{tab:system_pairise}
\end{table*}

% Results for AR: r-BLEU when aggregated at sentence-level gives
% (EN) 8, (BR-PT) 6, (IT) 5, and (FR) 5 -> (ALL) 24/40 correct pairwise 
% comparisons. 

\section{Evaluated Systems Details}\label{sec:evaluated_systems}

For each of \portuguese, \italian, and \french, 
outputs are sampled from:

\begin{enumerate}
    \item Rule-based systems consisting of hand-crafted transformations (e.g., fixing casing, normalizing punctuation, expanding contractions, etc.);
    \item Round-trip translation models that pivot to \english and backtranslate to the original language;
    \item Bi-directional neural machine translation~(\textsc{mt}) models that employ side constraints to perform style transfer for both directions of formality (i.e., informal$\leftrightarrow$formal)---trained  on (machine) translated informal-formal pairs of an English parallel corpus (i.e., \textsc{gyafc});
    \item Bi-directional \textsc{nmt} models that augment the training data of 3. via backtranslation of informal sentences;
    \item A multi-task variant of 3. that augments the training data with parallel-sentences from bilingual resources (i.e., OpenSubtitles) and learns to translate jointly between and across languages.
\end{enumerate}
For \english, the outputs were sampled from:
\begin{enumerate}
    \item A rule-based system of similar transformations to ones for \portuguese, \french, and \italian;
    \item A phrase-based machine translation model trained on informal-formal pairs of \textsc{gyafc};
    \item An \textsc{nmt} model trained on \textsc{gyafc} to perform style transfer uni-directionaly;
    \item A variant of 3. that incorporates a copy-enriched mechanism that enables direct copying of words from input;
    \item A variant of 4. trained on additional back-translated data of target style sentences using 2. 
\end{enumerate}

In general, neural models performed best for all languages according to overall human judgments, while
the simpler baselines perform closer to the more advanced neural models for \portuguese, \french, and \italian.
For each evaluation dimension  $500$ outputs are evaluated for \english and $100$ outputs per system for \portuguese, \french, and \italian.

\section{Meaning Preservation Metrics (reference-based)}
Table~\ref{tab:meaning_preservation_metrics_ref_based_correlations_with_human} presents supplemental results on meaning preservation metrics for reference-based settings.
\begin{table}[!ht]
    \centering
    \scalebox{0.75}{
    \begin{tabular}{lrrrr}
    \rowcolor{gray!10}
    \textbf{\textsc{metric}}     & \textbf{\textsc{en}} & \textbf{\textsc{it}} & \textbf{\textsc{pt}} & \textbf{\textsc{fr}} \\
    r-\textsc{bleu} & 0.306 &0.004 &0.047 &0.122 \\
    \textsc{meteor} & 0.279 &-0.005 &0.061 &0.124 \\
    chr\textsc{f} & 0.319 &0.065 &0.044 &0.174 \\
    \textsc{wmd} & 0.316 &0.039 &0.098 &0.198 \\
    Cosine & 0.218 &0.027 &0.048 &0.161 \\
    \textsc{bert}-score &  0.359 &-0.023 &0.054 &0.112 \\
    m\textsc{bert} (\textsc{translate-train}) & 0.369 &0.077 &0.167 &0.165 \\
    m\textsc{bert} (\textsc{zero-shot}) & - &0.124 &\textbf{0.197} &0.179 \\
    \textsc{xlm-r} (\textsc{translate-train}) & \textbf{0.385} &\textbf{0.183} &0.136 &\textbf{0.259} \\
    \textsc{xlm-r} (\textsc{zero-shot}) & - &0.179 &0.153 &0.258 \\
    \end{tabular}}
    \caption{Spearman's $\rho$ correlation of meaning preservation metrics for reference-based meaning.}
    \label{tab:meaning_preservation_metrics_ref_based_correlations_with_human}
\end{table}

% Remember to comment this out!
%\input{tables/cross-metric}

\end{document}